 \documentclass[final,5p,times,twocolumn,numeric,dashed=false]{elsarticle}

\usepackage{stfloats}
\usepackage{float}
\usepackage{amssymb}
\usepackage{natbib}
\usepackage{booktabs}
\usepackage{xurl}
\usepackage{todonotes}
\usepackage{caption}
\usepackage{subcaption}
\usepackage{graphicx}
\usepackage{amsmath}
\usepackage{dirtytalk}
\usepackage{xtab}
\usepackage{xtab,afterpage}
\usepackage{longtable}

\usepackage{array}
\newcolumntype{x}[1]{>{\centering\arraybackslash\hspace{0pt}}p{#1}}

\newcolumntype{y}[1]{>{\raggedright\arraybackslash\hspace{0pt}}p{#1}}

\journal{arxiv}

\begin{document}

\UseRawInputEncoding

\begin{frontmatter}



\title{Detecting Throat Cancer from Speech Signals using Machine Learning: A Scoping Literature Review}


\author[first]{Mary Paterson \corref{cor1}}
\affiliation[first]{organization={Faculty of Engineering and Physical Sciences University of Leeds},
            city={Leeds},
            country={UK}}

\author[second]{James Moor}
\affiliation[second]{organization={Ear, Nose and Throat Department Leeds Teaching Hospitals NHS Trust},
            city={Leeds},
            country={UK}}

\author[first]{Luisa Cutillo}

\cortext[cor1]{Corresponding author: scmlp@leeds.ac.uk}

\begin{abstract}
{\textit{Introduction:} Cases of throat cancer are rising worldwide. With survival decreasing significantly at later stages, early detection is vital. Artificial intelligence (AI) and machine learning (ML) have the potential to detect throat cancer from patient speech, facilitating earlier diagnosis and reducing the burden on overstretched healthcare systems. However, no comprehensive review has explored the use of AI and ML for detecting throat cancer from speech. This review aims to fill this gap by evaluating how these technologies perform and identifying issues that need to be addressed in future research. 

\textit{Materials and Methods:} We conducted a scoping literature review across three databases: Scopus, Web of Science, and PubMed. We included articles that classified speech using machine learning and specified the inclusion of throat cancer patients in their data. Articles were categorized based on whether they performed binary or multi-class classification.

\textit{Results:} We found 27 articles fitting our inclusion criteria, 12 performing binary classification, 13 performing multi-class classification, and two that do both binary and multiclass classification. The most common classification method used was neural networks, and the most frequently extracted feature was mel-spectrograms. We also documented pre-processing methods and classifier performance. We compared each article against the TRIPOD-AI checklist, which showed a significant lack of open science, with only one article sharing code and only three using open-access data. 

\textit{Conclusion:} Open-source code is essential for external validation and further development in this field. Our review indicates that no single method or specific feature consistently outperforms others in detecting throat cancer from speech. Future research should focus on standardizing methodologies and improving the reproducibility of results. }

\end{abstract}




\end{frontmatter}




\section{Introduction}
\label{introduction}
\subsection{Background}
The World Health Organisation predicts that cases of throat cancer (cancer of the larynx, nasopharynx, oropharynx, and hypopharynx) will rise by 15\% by 2040 \cite{ferlay_global_2020}. Early diagnosis is key to successful treatment, with Cancer Research UK stating that approximately 90\% of patients will survive for five years or more after diagnosis with stage one laryngeal cancer compared to only 30\% of patients diagnosed with stage four \cite{cancer_research_uk_survival_2019}. Late diagnosis also changes treatment options, with later-stage cancers requiring more aggressive and invasive treatment than early-stage cancers \cite{cancer_research_uk_treatment_2021, jones_laryngeal_2016}. Head and neck cancers, including throat cancer, are more common in individuals in low socioeconomic groups; as such, it's important that diagnostic tools are developed to reduce the cost of diagnosis \cite{lins_socio-demographic_2019}. 

Upon presentation with symptoms, patients undergo diagnostic techniques such as clinical history assessment, voice evaluation by clinicians, nasendoscopy, laryngoscopy, and biopsy \cite{nhs_laryngeal_2017}. Nasendoscopy, a standard outpatient procedure, uses a fibre-optic endoscope inserted through the nose to view the larynx and hypopharynx. If abnormalities are detected, biopsies can be taken under local anaesthesia in an outpatient setting or during a laryngoscopy under general anaesthesia for a comprehensive assessment of the larynx \cite{nhs_laryngeal_2017}.

Identifying cancers is challenging and highly dependent on timely symptom recognition. Common early symptoms of throat cancer include changes in voice, pain or difficulty swallowing, and lumps in the neck  \cite{nhs_laryngeal_2018}. There are many benign voice disorders that have overlapping symptoms with throat cancer, making them difficult to differentiate without these invasive tests \cite{mayo_clinic_vocal_2022, mayo_clinic_voice_2022}. Once cancer has been ruled out, speech assessment can be used to assist in the diagnosis of voice disorders and to measure the degree of abnormality. Two main protocols are commonly used: GRBAS and the Consensus Auditory-Perceptual Evaluation of Voice (CAPE-V). GRBAS is a reproducible clinical assessment of abnormal voices commonly used in clinical practice. Five aspects of the patient’s voice (grade, roughness, breathiness, asthenia, and strain) are scored on a scale of zero to three, where zero is normal, one is mild, two is moderate, and three is severe \cite{reghunathan_components_2019}. CAPE-V is an assessment tool which requires patients to perform three speech tasks, prolonged vowel sounds, reading sentences aloud, and spontaneous speech. Six speech features are assessed for each task - overall severity, roughness, breathiness, strain, pitch, and loudness. Each feature is rated on a scale from 0 (normal) to 100 (severe) \cite{kempster_consensus_2009}. These protocols are subjective and require expert knowledge.

With healthcare systems being under immense pressure, it's important that resources are correctly and most efficiently allocated. In the UK, only 2.7-4.3\% of patients referred on the urgent suspected cancer pathway for head and neck cancers had a cancer diagnosis; if tools are able to detect non-cancer patients for more efficient referral, some of the load on healthcare systems could be eased \cite{nhs_digital_urgent_2024}. 

Since a hoarse voice is one of the most common symptoms of voice disorders, including cancers, detecting throat cancer from speech using Artificial intelligence (AI) and machine learning (ML) has been suggested as an alternative to the invasive and expensive procedures currently used for the diagnosis of throat cancer.  AI and ML do not have a single definition. \citeauthor{ibm_what_2023} \cite{ibm_what_2023} defines AI as computers simulating human intelligence. \citeauthor{russell_artificial_2010} \cite{russell_artificial_2010} state that AI should behave rationally and human-like. \citeauthor{dstl_artificial_2020} \cite{dstl_artificial_2020} define AI as \say{Theories and techniques developed to allow computer systems to perform tasks normally requiring human or biological intelligence} and ML as \say{A field that aims to provide computer systems with the ability to learn and automatically improve without having to be explicitly programmed}. In this work, we define AI and ML similarly and accept works as having used AI or ML if they use a computer program that was not explicitly programmed but learned from data, excluding any simple statistical methods.

It has been suggested that by using ML or AI, throat cancer may be able to be detected from patient speech \cite{kim_convolutional_2020, wang_detection_2022, kwon_diagnosis_2022}. This technology may be able to detect throat cancer sooner and help prioritise those patients who are most at risk while reducing the need for invasive and expensive diagnostic procedures. In this work, we perform a scoping literature review to document and analyse current ML and AI methods for the detection of throat cancer from speech. 


\subsection{Related Work}
Current work in the area of using patient speech for pathology detection using AI generally focuses on neurological conditions, most commonly dementia and Parkinson's disease \cite{al-hameed_new_2019, elen_comparison_2020}. There have been several conference challenges focused on the detection of dementia from speech \cite{luz_alzheimers_2020, luz_detecting_2021,luz_multilingual_2023}. In these challenges, participants were tasked with classifying patients with Alzheimer's dementia from non-Alzheimer's patients. These challenges use a mixture of acoustic and linguistic features in classification. This area differs significantly from the detection of throat cancer, as neurological diseases often lead to a decline in linguistic ability \cite{cummins_comparison_2020}. However, since throat cancer is a structural disorder rather than neurological, there is no effect on a patient's linguistic ability but only on the acoustic features of speech. 

\subsection{Objective} \label{sec:RQs}
The objective of this literature review is to explore current work using AI and ML for detecting throat cancer from speech. We formulated the following research questions (RQ):

\begin{itemize}
    \item RQ1: What machine learning and artificial intelligence methods have been used for the detection of vocal pathologies, including throat cancer, from patient speech?
    \item RQ2: What features of speech can be used to identify pathological speech, including throat cancer?
    \item RQ3: What are the strengths of the existing research, and what issues need to be addressed in future work?
\end{itemize}

\section{Methods}
\subsection{Search Strategy}

This literature search was conducted across three databases: Scopus, Web of Science, and PubMed and was run on 14/10/2024. The publication's title, abstract, and keywords were searched using the terms shown in Table \ref{tab:SearchTerms}. These search terms were chosen to try to cover all topics that should be included in articles relevant to this literature search. These topics are cancer, throat, speech, detection, and machine learning. The terms used for cancer, throat, and speech were discussed with an experienced ENT consultant and the terms used for detection and machine learning were discussed with a technical expert to ensure that all relevant terms were included.

\begin{table*}[htbp]
    \centering
    \begin{tabular}{lllll}
    \toprule
			Cancer & Throat & Speech & Detection & Machine Learning \\
			\midrule 
			cancer* & throat & audio & classifi* & ``machine learning"  \\
			carcinoma* & larynx & speech & diagnos* & ``artificial intelligence"  \\
			tumour* & laryngeal & sound & detection & ``neural networks" \\ 
			tumor* & ``voice box" & spectrograms &  & ``neural network"\\
			neoplasm* & glottis & voice &  & ``deep learning"\\ 
			malignan* & glottic & vocal &  & ml \\ 
        	 & ``vocal cord" & prosody &  & ai \\  		
        	 & supraglottis & acoustic &  & \\		
        	 & supraglottic  &  &  & \\	
              & pharynx & & & \\
			\bottomrule 
    \end{tabular}
    \caption{The search terms used in the literature review. Terms within `` " are found exactly not separated, a * in a term can be replaced by any characters (for example, classifi* can be classifi\textit{cation}, classi\textit{fier}, classi\textit{fiers} etc).}
    \label{tab:SearchTerms}
\end{table*}

\subsection{Inclusion and Exclusion Criteria}
In this review we include research study designs and exclude review articles, case studies, and clinical trials. Articles were also excluded if they were not written in English and if a full version of the manuscript was not available. No restrictions were placed on publication dates.

To be included in this study, the articles had to adhere to the following inclusion criteria.
\begin{itemize}
    \item Specifically state that patients with throat cancer are included in their dataset.
    \item Use machine learning or artificial intelligence methods to classify patients with a vocal pathology.
    \item Use speech recordings (or features obtained from speech) as the primary input to the classification system.
\end{itemize}

\subsection{Study Selection}
\begin{figure*}
    \centering
    \includegraphics[width=0.8\textwidth]{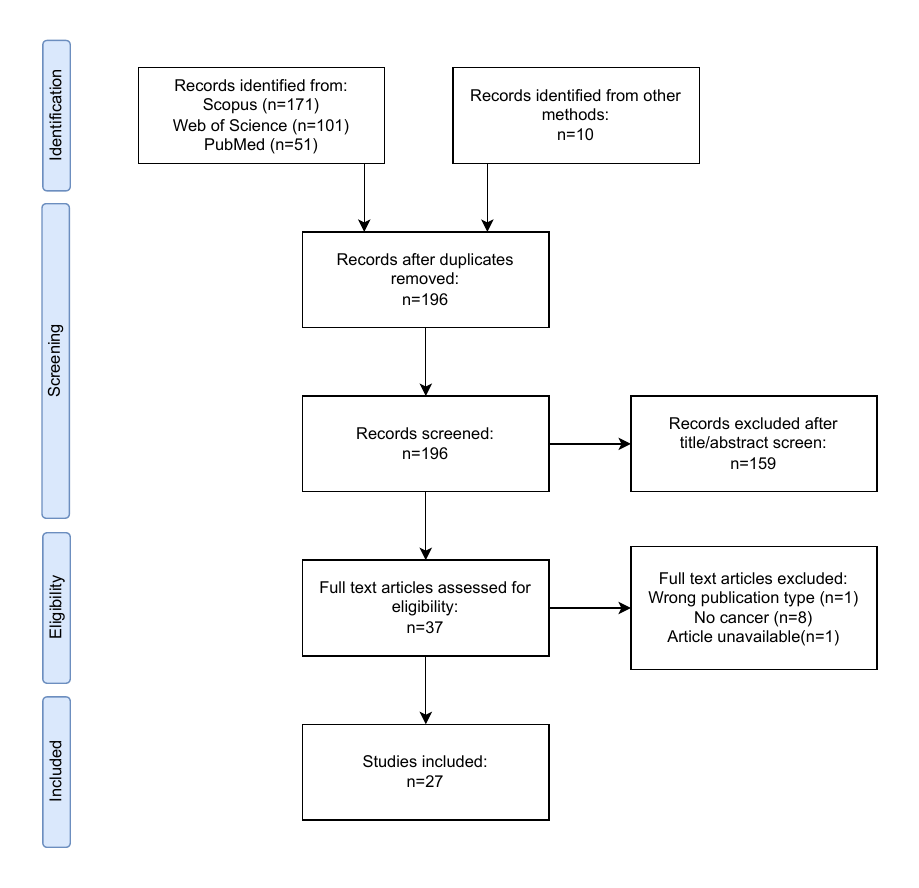}
    \caption{The Prisma diagram shows the steps taken to obtain relevant articles for this literature search.}
    \label{fig:Prisma}
\end{figure*}

Using the search terms stated in Table \ref{tab:SearchTerms} across Scopus, Web of Science, and PubMed, a total of 323 articles were found, and ten additional articles were also identified as part of a relevant challenge. All articles were imported into Zotero for reference management \cite{corporation_for_digital_scholarship_zotero_2024}. Rayyan was used to facilitate the removal of duplicates (n=127) as well as being used for screening \cite{ouzzani_rayyanweb_2016}. All article titles and abstracts were screened, and ineligible articles were excluded (n=159). We then conducted a full article screening and removed any ineligible or inaccessible articles (n=10), resulting in 27 articles being included. Of the articles excluded at full-text screening, the majority (n=8) were excluded because they did not explicitly state the inclusion of cancer patients in their dataset, often looking to precancerous conditions instead. One article was the wrong study type, and one article was unavailable. Figure \ref{fig:Prisma} shows a flowchart based on the PRISMA framework depicting the process used to identify and screen the articles found in the search \cite{page_prisma_2021}. Two researchers carried out the screening process, and any conflicts were resolved through discussion.

\subsection{Data Extraction}
\begin{figure*}[htbp]
    \centering
    \includegraphics[width=0.8\textwidth]{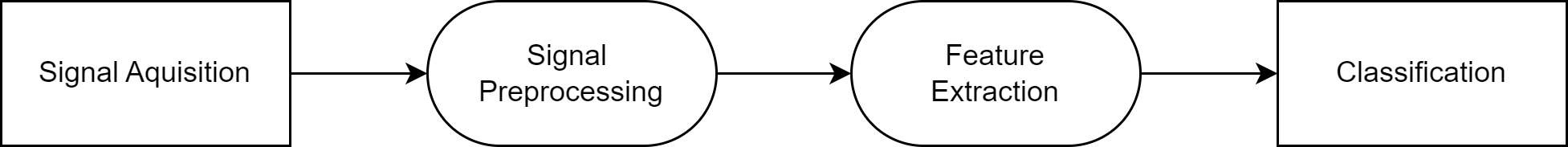}
    \caption{The typical pipeline for classifying pathological speech using machine learning. Both preprocessing and feature extraction are optional steps not performed by all articles.}
    \label{fig:ClassPipeline}
\end{figure*}

Work in this area follows a typical workflow shown in Figure \ref{fig:ClassPipeline}.  Preprocessing steps are any processing on the raw signal prior to feature extraction or classification. Feature extraction is commonly performed prior to classification and involves manipulating the signal in some way to extract features; these features are often meaningful and reduce the input size into the classification system, although some works may classify the signal without extracting any features. 

A single reviewer extracted the following data from each article:
\begin{itemize}
    \item The source and full reference.
    \item Any preprocessing steps performed.
    \item The features used in classification.
    \item The classification methods used.
    \item The results obtained.
\end{itemize}

The results of all attempted models were extracted from each article. The most common metrics across all articles were accuracy, specificity, and sensitivity. We show the equations for these metrics below:

\begin{equation} \label{eq:Accuracy}
\text{Accuracy} = \frac{\text{Correctly classified speech}}{\text{Total speech samples}}
\end{equation}

\begin{equation} \label{eq:Specificity}
\text{Specificity} = \frac{\text{Correctly classified healthy speech}}{\text{Total healthy speech}}
\end{equation}

\begin{equation} \label{eq:Sensitivity}
\text{Sensitivity} = \frac{\text{Correctly classified pathological speech}}{\text{Total pathological speech}}
\end{equation}

Another metric often used in the papers using multi-class classification was unweighted average recall (UAR). This metric is calculated by averaging the recall value for each of the specific pathologies included in the dataset. Equation \ref{eq:UAR} shows how it is calculated, where N is the number of pathologies in the dataset and $R_i$ is the recall (Equation \ref{eq:Recall}) of the ith pathology in the dataset.

\begin{equation}\label{eq:Recall}
    R_i = \frac{\text{Correctly classified speech of pathology i}}{\text{Total number of pathology i speech samples}}
\end{equation}

\begin{equation} \label{eq:UAR}
    \text{UAR} = \frac{\sum_{i=1}^{N}{R_i}}{N}
\end{equation}

\section{Results and Discussion}
\subsection{Overview} \label{sec:Overview}
 In this section, we provide a detailed overview of the articles included in this study. We grouped the articles based on whether they performed binary or multi-class classification. The extracted data from each article can be seen in Tables \ref{tab:BinaryResults} and \ref{tab:MultiResults} for articles performing binary and multi-class classification, respectively. In both tables, we give the best performance reported in each article in terms of accuracy (Equation \ref{eq:Accuracy}) and the corresponding specificity (Equation \ref{eq:Specificity}), and sensitivity (Equation \ref{eq:Sensitivity}), as those were the most commonly reported metrics. In Table \ref{tab:MultiResults} We also report the UAR (Equation \ref{eq:UAR}) for the multi-classification articles, as this was most commonly reported in these articles. 

 Of the 27 articles included in this review 11 articles perform binary classification, 14 articles perform multi-class classification, and two articles perform both binary and multi-class classification. Additionally, we note that one of the articles reviewed is our previous work; despite this, we aim to maintain objectivity in this review \cite{paterson_pipeline_2023}. 

\begin{table*}
\begin{center}
\resizebox{\textwidth}{!}{%
\begin{tabular}{|y{2.5cm}|y{2.5cm}|y{2.5cm}|y{2.5cm}|l|l|l|}
\hline
Paper & Preprocessing & Features & Classification Method(s) & Accuracy & Sensitivity & Specificity \\ \hline
\citeauthor{gavidia-ceballos_direct_1996} \cite{gavidia-ceballos_direct_1996} & Wiener filter & Spectral features\newline Acoustic features & HMM & 82.9 & X & X \\ \hline
\citeauthor{godino-llorente_automatic_2004} \cite{godino-llorente_automatic_2004} & Antialiasing filter\newline Windowing\newline Endpoint detection & MFCC & MLP \newline LVQ & 96 & 95.67 & 96.73 \\ \hline
\citeauthor{ben_aicha_cancer_2016} \cite{ben_aicha_cancer_2016} & X & Glottal Parameters & ANN & 96.9 & X & X \\ \hline
\citeauthor{ezzine_towards_2016} \cite{ezzine_towards_2016} & X & Glottal Parameters & MLP & 98 & X & X \\ \hline
\citeauthor{fang_detection_2019} \cite{fang_detection_2019} & X & MFCC & DNN \newline GMM \newline SVM & 99.14 & X & X \\ \hline
\citeauthor{kim_convolutional_2020} \cite{kim_convolutional_2020} & Normalization & Acoustic features\newline MFCC \newline STFT \newline Raw signals & SVM \newline XGBoost \newline LightGBM \newline ANN \newline CNN (1D and 2D) & 85.2 & 78.0 & 93.3 \\ \hline
\citeauthor{kwon_diagnosis_2022} \cite{kwon_diagnosis_2022} & X & MFCC & CNN \newline DT & 87.88 & 94.12 & 81.25 \\ \hline
\citeauthor{wang_detection_2022} \cite{wang_detection_2022} & X & MFCC & DNN & 86.11 & 77.78 & 88.89 \\ \hline
\citeauthor{chen_classification_2023} \cite{chen_classification_2023} & X & MFCC \newline Chroma vectors & SVM & 85.00 & X & X \\ \hline
\citeauthor{paterson_pipeline_2023} \cite{paterson_pipeline_2023} & Band-pass filter \newline Wiener filter \newline Wavelet filter \newline Endpoint detection \newline Normalization & Acoustic features & SVM \newline DT \newline RF \newline LR & 81.2 & 87.2 & X \\ \hline
\citeauthor{zaim_accuracy_2023} \cite{zaim_accuracy_2023} & X & MFCC & OSLEM \newline NB \newline SVM \newline DT & 92 & 100 & 58 \\ \hline
\citeauthor{kim_classification_2024} \cite{kim_classification_2024} & X & MFCC & SVM \newline LightGBM \newline ANN & 96.51 & 93.94 & X \\ \hline
\citeauthor{wang_ai_2024} \cite{wang_ai_2024} & X & Raw signals \newline Medical records & CNN & 81.5 & 78.3 & 81.6 \\ \hline
\end{tabular}%
}

\caption{A summary of the papers using binary classification. The articles are presented in chronological order. MFCC - Mel-frequency cepstral coefficients, STFT - short-term Fourier transform, SVM - support vector machine, XGBoost - extreme gradient boosting, LightGBM - light gradient-boosting machine, ANN - artificial neural network, CNN - convolutional neural network, DT - decision tree, RF - random forest, DNN - deep neural network, MLP - multi-layered perceptron, LVQ - learning vector quantization, GMM - Gaussian mixture models, HMM - hidden Markov model, OSLEM - Online Sequential Extreme Learning Machine, NB - Naive Bayes, LR - logistic regression}

\label{tab:BinaryResults}
\end{center}
\end{table*}

\onecolumn

\begin{longtable}{|y{2.5cm}|y{2cm}|y{2.5cm}|y{2.5cm}|l|l|l|l|}
\hline
Paper & Preprocessing & Features & Classification Method(s) & Accuracy & Sensitivity & Specificity & UAR\\ 
 \hline
 \endfirsthead

 \hline
 \multicolumn{8}{|c|}{Continuation of Table \ref{tab:MultiResults}}\\
 \hline
Paper & Preprocessing & Features & Classification Method(s) & Accuracy & Sensitivity & Specificity & UAR \\
 \hline
 \endhead
 
 \multicolumn{8}{|c|}{Table \ref{tab:MultiResults}: A summary of the papers using multi-class classification \textit{(continued below)}.} \\
 \hline
 \endfoot

\caption{A summary of the papers using multi-class classification. The articles are presented in chronological order. MFCC - Mel-frequency cepstral coefficients, DNN - deep neural network, NN - neural network, SVM - support vector machine, MIL - multi-instance learning, GMM - Gaussian mixture model, GBT - gradient boosting tree, k-nn - k nearest neighbour, RF - random forest, LR - linear regression, CNN - convolutional neural network, LSTM-FCN - long short term memory fully convolutional network}\label{tab:MultiResults}

 \endlastfoot

\citeauthor{verikas_towards_2010} \cite{verikas_towards_2010} & X & Medical records \newline Acoustic features \newline Cosine transform coefficients & SVM & 80.47 & X & X & X \\ \hline
\citeauthor{arias-londono_byovoz_2018} \cite{arias-londono_byovoz_2018} & Voice activity detection \newline Normalization & Perturbation features\newline MFCC \newline Spectrum features \newline Complexity features & GMM \newline GBT\newline SVM\newline k-nn\newline RF & X & 92 & 54 & 61 \\ \hline
\citeauthor{bhat_femh_2018} \cite{bhat_femh_2018} & X & Acoustic features \newline MFCC \newline Spectral features & BayesNet \newline RF & X & 96.6 & 66 & 68.67 \\ \hline
\citeauthor{chuang_dnn-based_2018} \cite{chuang_dnn-based_2018} & X & MFCC & DNN & X & 93.1 & 46 & 62.87 \\ \hline
\citeauthor{degila_ucd_2018} \cite{degila_ucd_2018} & X & MFCC \newline Spectral features & SVM \newline RF \newline NN\newline LR & X & 89.4 & 54.0 & 71.30 \\ \hline
\citeauthor{grzywalski_parameterization_2018} \cite{grzywalski_parameterization_2018} & Low-pass filtering\newline Cropping\newline Data augmentation & MFCC \newline Acoustic features & DNN & X & 89.4 & 66.0 & 71.2 \\ \hline
\citeauthor{islam_transfer_2018} \cite{islam_transfer_2018} & Segmentation & Prosodic features \newline MFCC\newline Acoustic features & SVM & X & 94.9 & 20 & 59.77 \\ \hline
\citeauthor{ju_multi-representation_2018} \cite{ju_multi-representation_2018} & Endpoint detection & Acoustic features \newline MFCC \newline Spectral features & MIL \newline SVM \newline Label propagation \newline Transductive SVM & X & X & X & 60.67 \\ \hline
\citeauthor{pishgar_pathological_2018} \cite{pishgar_pathological_2018} & X & MFCC & SVM \newline XGBoost \newline LSTM-FCN & X & 88.60 & 78.23 & 59.00 \\ \hline
\citeauthor{pham_diagnosing_2018} \cite{pham_diagnosing_2018} & Endpoint detection & MFCC & SVM \newline RF\newline k-nn \newline Gradient Boosting & 68.48 & X & X & X \\ \hline
\citeauthor{ramalingam_ieee_2018} \cite{ramalingam_ieee_2018} & X & Spectral features \newline MFCC \newline Time domain features & CNN \newline RNN & 93 & 96 & 18 & X \\ \hline
\citeauthor{fang_combining_2019} \cite{fang_combining_2019} & X & Medical records\newline MFCC & DNN \newline GMM & 87.26 & X & X & 81.59 \\ \hline
\citeauthor{miliaresi_combining_2021} \cite{miliaresi_combining_2021} & Normalization \newline Segmentation & MFCC\newline Acoustic features \newline Medical records & NN & 57 & X & X & X \\ \hline
\citeauthor{chen_classification_2023} \cite{chen_classification_2023} & X & MFCC \newline Chroma vectors \newline Mel-spectrogram & SVM \newline ImageNet & 51.38 & X & X & X \\ \hline
\citeauthor{song_enhancing_2023} \cite{song_enhancing_2023} & X & MFCC \newline Acoustic features \newline Demographic data & ResNet & 99.69 & 100 & 100 & X \\ \hline
\citeauthor{kim_classification_2024} \cite{kim_classification_2024} & X & MFCC & SVM \newline LightGBM \newline ANN & 82.61 & 72.94 & X & X \\ \hline

\end{longtable}

\twocolumn
\noindent

Our review covers articles published between 1996 and 2024, with no date restrictions applied. The distribution of publication years shows a peak of 10 articles published in 2018 due to a challenge that year and an increasing trend in the last three years (2022-2024) with eight articles published. The majority of these articles were conference papers (n=16) compared to journal articles (n=11). However, we do note that the majority of the conference articles (n=10) are from a single challenge.

Tables \ref{tab:BinaryResults} and \ref{tab:MultiResults} show the range of results achieved in the articles found during this search. In Table \ref{tab:BinaryResults} accuracy ranges from 80-99\%, in Table \ref{tab:MultiResults} accuracy ranges from 50-99\% with UAR ranging from 59-81\%. There are two articles of note in these tables which achieve exceptional results. In Table \ref{tab:BinaryResults} it can be seen that \citeauthor{fang_detection_2019} \cite{fang_detection_2019} achieves an accuracy score of 99.14\% on a binary classification task. This, however, was achieved on an external validation which, upon further inspection, does not contain any cancer patients \cite{fan_class-imbalanced_2021}. The best result on the cross-validation was  94.26\% on male patients and 90.52\% on female patients.

In Table \ref{tab:MultiResults}, we see that \citeauthor{song_enhancing_2023} \cite{song_enhancing_2023} has exceptional results when compared to the other articles. However, there is a discrepancy in the number of patients in the dataset for each class and the number reported in the confusion matrix (257 patients in the dataset, 977 samples in the confusion matrix). It is unclear from the article why these numbers differ; it is possible that patients provided multiple recordings or audio splitting was conducted. The results presented are calculated from five-fold cross-validation, and as such, we hypothesise that data leakage may have occurred if care had not been taken to ensure that multiple recordings from the same patient had not been split across the validation folds. 

During data extraction, all results presented in each article were extracted. The majority of articles present more than one result, with only three articles presenting one result \cite{chuang_dnn-based_2018, bhat_femh_2018, arias-londono_byovoz_2018}. \citeauthor{chen_classification_2023} \cite{chen_classification_2023} presents the most results with 44 models being evaluated. The average number of results presented is nine.  

\begin{figure}[htbp]
    \centering
    \includegraphics[width=\linewidth]{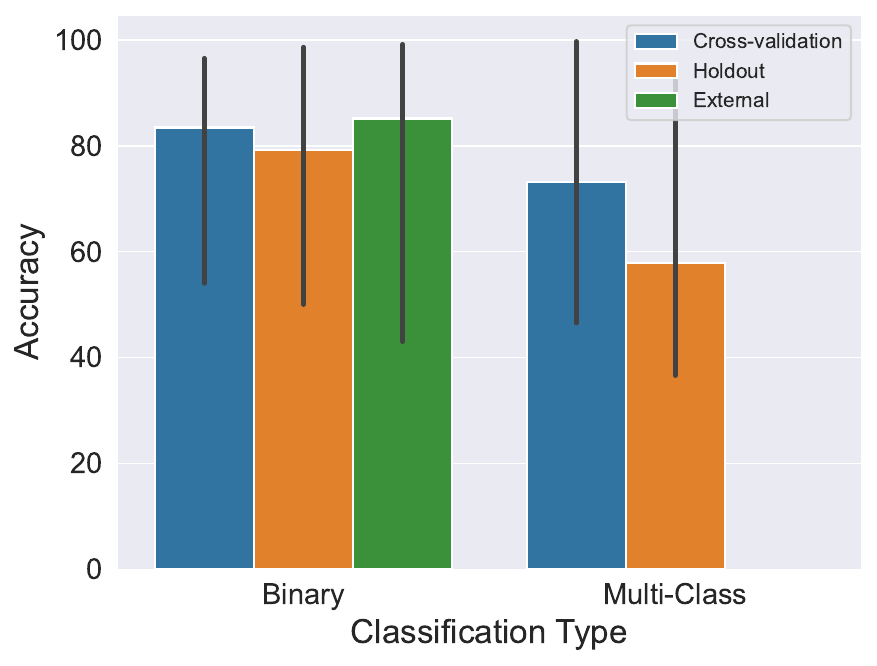}
    \caption{A comparison of the results obtained when evaluating using cross-validation, holdout, and external test sets. The bar represents the average accuracy, and the error bars are the minimum and maximum values.}
    \label{fig:EvaluationCompare}
\end{figure}

We also note that the articles are not consistent in their evaluation methods. Most articles used a single evaluation method, with ten only evaluating on a holdout test set and eight evaluating only on cross-validation. Some articles however, use a mix of evaluation methods, four articles evaluate using cross-validation and a holdout test set, two use cross-validation and an external test set, and one uses cross-validtaion, holdout, and external test sets. It was not clear how the models were evaluated in two of the articles \cite{gavidia-ceballos_direct_1996, ben_aicha_cancer_2016}. Figure \ref{fig:EvaluationCompare} shows the performance difference between the models evaluated on cross-validation, holdout, and external test sets. All results presented in the articles are considered when creating this figure. It can be seen that when evaluation is performed on cross-validation, the average performance is higher than in articles where evaluation is performed on a holdout test set. The results on external test sets are higher than expected. However, this is largely affected by the results presented by \citeauthor{fang_detection_2019} \cite{fang_detection_2019} as discussed above, this may be skewed due to discrepancies in the pathologies present in the external dataset. 

\subsubsection{Datasets}
Gathering real-world datasets is hard due to restrictions related to data privacy and the infrequent routine collection of relevant speech data. When datasets are collected, they are often small, including fewer than 50 cancer patients. 

Table \ref{tab:DatasetSummary} shows the size of the dataset used in each article splitting the datasets into the number of healthy, cancer, and non-cancer pathologies reported by each article. We found that some articles combine cancerous and precancerous conditions, where the number of cancerous and precancerous conditions is made clear, the number of precancerous samples is added to the ``Non-Cancer Pathology(s)" column. 

\citeauthor{ben_aicha_cancer_2016} \cite{ben_aicha_cancer_2016} are unclear in the number of patients in their dataset, simply stating that there are over 100 healthy participants without giving any specific count. There is a significant outlier in the number of samples used in the articles as \citeauthor{ezzine_towards_2016} \cite{ezzine_towards_2016} reports over 6000 samples. We investigated this further and found that in this article, the open-source dataset Saarbruecken Voice Database is used \cite{manfred_putzer_saarbruecken_2007}. This dataset contains over 2000 samples ($\sim$2700), and as such, the samples referred to by \citeauthor{ezzine_towards_2016} cannot be directly from the  Saarbruecken Voice Database. We speculate that the authors may have augmented their dataset by splitting audio samples into smaller segments, though this is not explicitly clarified in the article. This discrepancy means that it is impossible to reproduce their results even with the available data.  

\begin{figure}[htbp]
    \centering
    \includegraphics[width=1\linewidth]{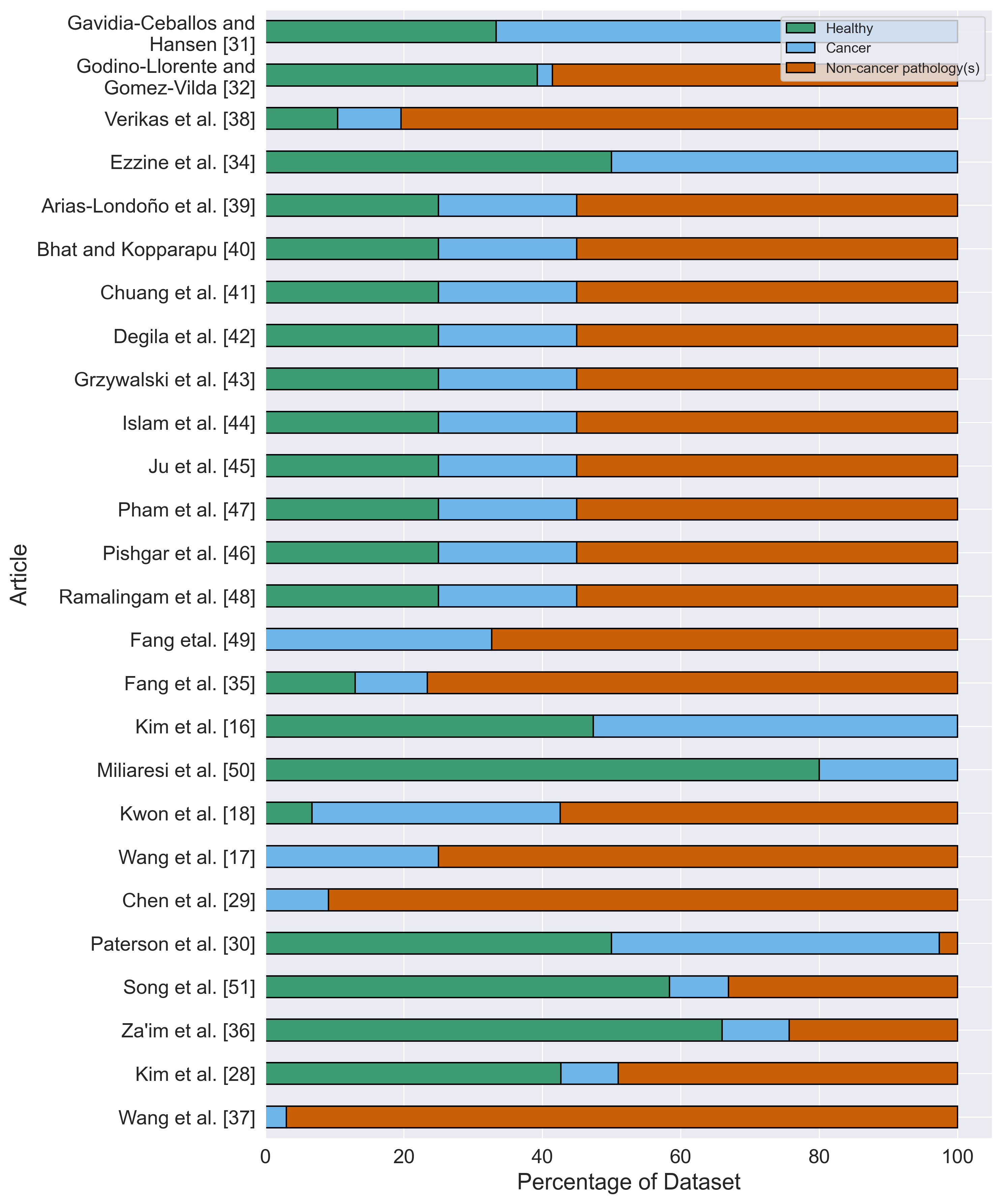}
    \caption{The percentage of samples for healthy, cancer, and non-cancer pathology(s) in each article. \cite{ben_aicha_cancer_2016} removed as exact counts were not provided.} 
    \label{fig:DatasetBreakdownPercentages}
\end{figure}

The next largest dataset is from \citeauthor{wang_ai_2024} \cite{wang_ai_2024}, with 2000 total patients. Figure \ref{fig:DatasetBreakdownPercentages} shows the percentage of patients that were healthy, had cancer and had non-cancer pathologies in each article. \citeauthor{gavidia-ceballos_direct_1996} \cite{gavidia-ceballos_direct_1996} has the highest percentage of cancer patients in their dataset (66.7\%) with \citeauthor{godino-llorente_automatic_2004} \cite{godino-llorente_automatic_2004} having the lowest percentage (2.2\%).

\begin{table}[htbp]
    \centering
    \begin{tabular}{y{3.3cm}|ccx{2cm}}
    \hline
        Article & Healthy  &Cancer & Non-Cancer Pathology(s) \\ 
        \hline
        \citeauthor{gavidia-ceballos_direct_1996}   \cite{gavidia-ceballos_direct_1996} & 10 & 20 & 0 \\
        \citeauthor{godino-llorente_automatic_2004}   \cite{godino-llorente_automatic_2004} & 53 & 3 & 79 \\
        \citeauthor{verikas_towards_2010}   \cite{verikas_towards_2010} & 25 & 22 & 193 \\
        \citeauthor{ben_aicha_cancer_2016}   \cite{ben_aicha_cancer_2016} & $>$100 & 101 & 0 \\
        \citeauthor{ezzine_towards_2016}   \cite{ezzine_towards_2016} & 3009 & 3009 & 0 \\
        \citeauthor{arias-londono_byovoz_2018}   \cite{arias-londono_byovoz_2018} & 50 & 40 & 110 \\
        \citeauthor{bhat_femh_2018}   \cite{bhat_femh_2018} & 50 & 40 & 110 \\
        \citeauthor{chuang_dnn-based_2018}   \cite{chuang_dnn-based_2018} & 50 & 40 & 110 \\
        \citeauthor{degila_ucd_2018}   \cite{degila_ucd_2018} & 50 & 40 & 110 \\
        \citeauthor{grzywalski_parameterization_2018}   \cite{grzywalski_parameterization_2018} & 50 & 40 & 110 \\
        \citeauthor{islam_transfer_2018}   \cite{islam_transfer_2018} & 50 & 40 & 110 \\
        \citeauthor{ju_multi-representation_2018}   \cite{ju_multi-representation_2018} & 50 & 40 & 110 \\
        \citeauthor{pham_diagnosing_2018}   \cite{pham_diagnosing_2018} & 50 & 40 & 110 \\
        \citeauthor{pishgar_pathological_2018}   \cite{pishgar_pathological_2018} & 50 & 40 & 110 \\
        \citeauthor{ramalingam_ieee_2018}   \cite{ramalingam_ieee_2018} & 50 & 40 & 110 \\
        \citeauthor{fang_combining_2019}   \cite{fang_combining_2019} & 0 & 84 & 173 \\
        \citeauthor{fang_detection_2019}   \cite{fang_detection_2019} & 60 & 48 & 354 \\
        \citeauthor{kim_convolutional_2020}   \cite{kim_convolutional_2020} & 45 & 50 & 0 \\
        \citeauthor{miliaresi_combining_2021}   \cite{miliaresi_combining_2021} & 200 & 50 & 0 \\
        \citeauthor{kwon_diagnosis_2022}   \cite{kwon_diagnosis_2022} & 33 & 176 & 282 \\
        \citeauthor{wang_detection_2022}   \cite{wang_detection_2022} & 0 & 43 & 129 \\
        \citeauthor{chen_classification_2023}   \cite{chen_classification_2023} & 0 & 40 & 398 \\
        \citeauthor{paterson_pipeline_2023}   \cite{paterson_pipeline_2023} & 38 & 36 & 2 \\
        \citeauthor{song_enhancing_2023} \cite{song_enhancing_2023} & 150 & 22 & 85 \\
        \citeauthor{zaim_accuracy_2023}   \cite{zaim_accuracy_2023} & 252 & 37 & 93 \\
        \citeauthor{kim_classification_2024}   \cite{kim_classification_2024} & 155 & 30 & 178 \\
        \citeauthor{wang_ai_2024}  \cite{wang_ai_2024} & 0 & 60 & 1940 \\
        
    \end{tabular}
    \caption{The reported samples in each article are split into healthy, cancer, and non-cancer pathologies.}
    \label{tab:DatasetSummary}
\end{table}

\subsubsection{Preprocessing}

Preprocessing is a critical step in data preparation for machine learning, aimed at enhancing the quality of data and improving model performance. In this study, we examined the preprocessing methods reported in the reviewed articles. We found that the majority of articles did not report any preprocessing steps (n=16). Of the articles performing binary classification, only four used preprocessing, each employing different methods. For multi-class classification, six articles used preprocessing, while nine did not. 

The most common preprocessing technique identified was endpoint detection (including cropping and voice activity detection), followed by normalization and filtering. Some articles did not specify the reasons for their choice of preprocessing methods, indicating that these choices may have been made on an ad-hoc basis rather than for specific reasons supported by the literature. Although \citeauthor{kim_convolutional_2020} \cite{kim_convolutional_2020} and \citeauthor{paterson_pipeline_2023} \cite{paterson_pipeline_2023} state that normalization reduces any effect that the distance between the microphone and the patient may have. \citeauthor{godino-llorente_automatic_2004} \cite{godino-llorente_automatic_2004} state that endpoint detection is used to avoid processing periods of silence, and \citeauthor{paterson_pipeline_2023} \cite{paterson_pipeline_2023} states that filtering is used for the removal of background noise.

\begin{figure}[htbp]
    \centering
    \includegraphics[width=\linewidth]{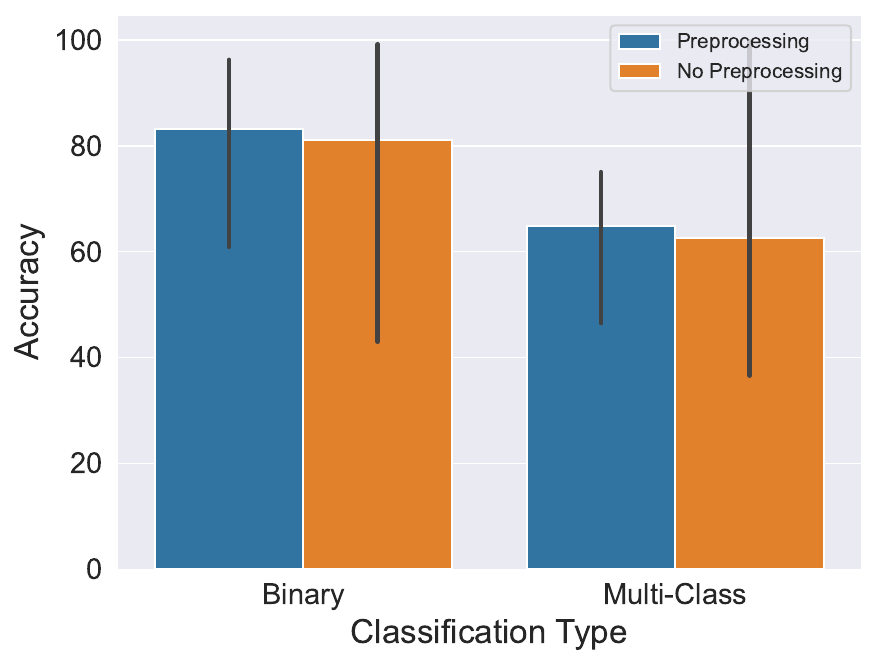}
    \caption{A comparison of the results obtained when signals were preprocessed and when they were not. The bar represents the average accuracy, and the error bars are the minimum and maximum values.}
    \label{fig:PreprocessingCompare}
\end{figure}

Figure \ref{fig:PreprocessingCompare} shows the difference in accuracy when signals were and weren't preprocessed. It can be seen that the mean values are similar. However, the minimum and maximum values are more extreme when signals are not preprocessed. This may indicate that preprocessing does improve model robustness. 

\subsection{RQ1 - Classification Methods}
In this section, we discuss Research Question 1: What machine learning and artificial intelligence methods have been used for the detection of vocal pathologies, including throat cancer, from patient speech? We present and evaluate the methods used in the examined articles and discuss their merits and drawbacks. 

\begin{figure}[htbp]
    \centering
    \includegraphics[width=0.5\textwidth]{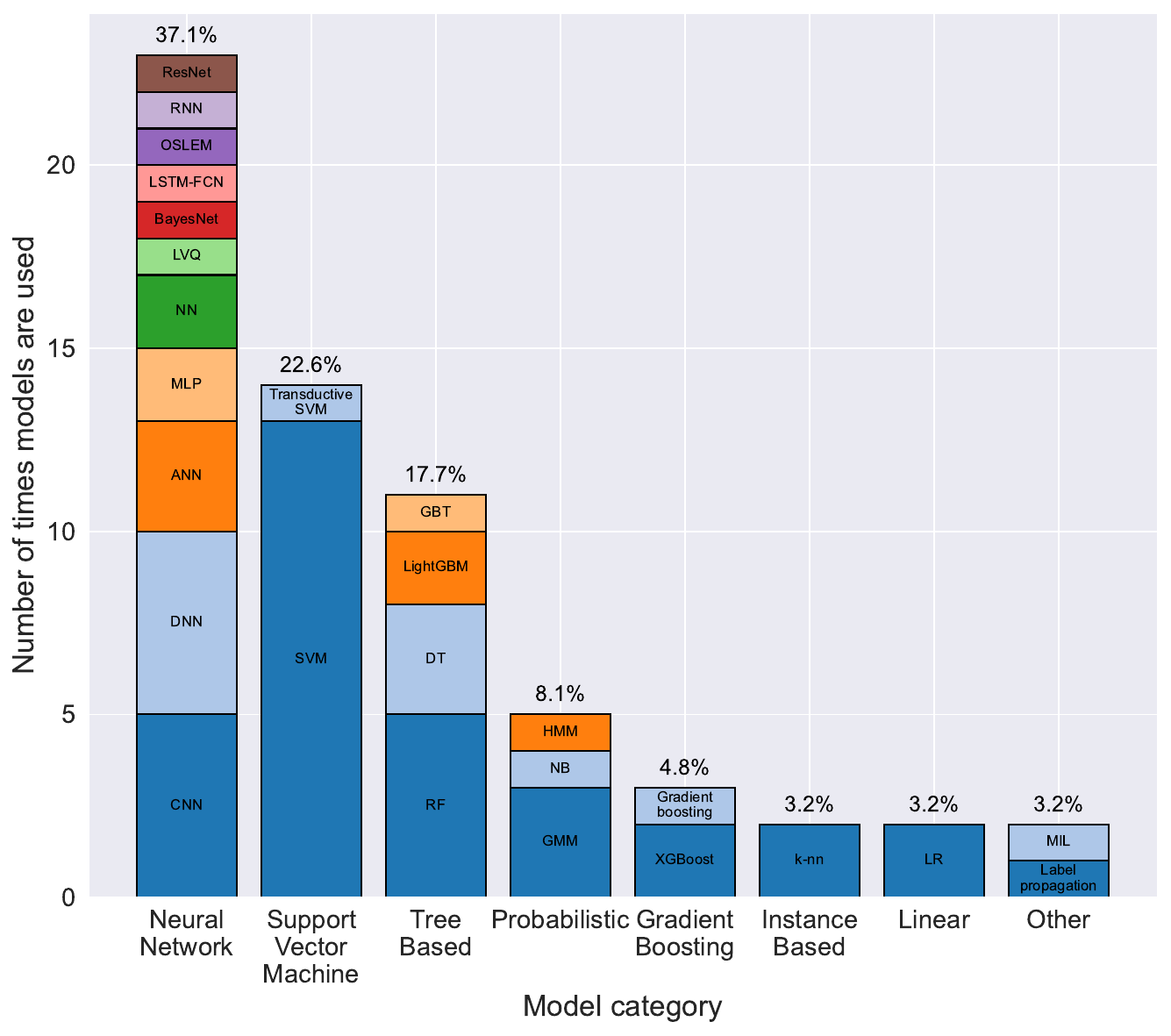}
    \caption{The classification methods used across all articles found in this review. Models have been categoriesed for ease of understanding.}
    \label{fig:ClassMethods}
\end{figure}

An average of 1.9 model types were used in each article with \citeauthor{pham_diagnosing_2018} \cite{pham_diagnosing_2018} using the most model types (SVM, random forest, k-nn, gradient boosting, and ensemble). Figure \ref{fig:ClassMethods} shows the breakdown of the classification methods used in the articles. For clarity, methods have been categorised, although the number of specific methods can still be seen in the stacked bar chat. The most common type of models used are neural networks, followed by support vector machines (SVM). 

Neural networks were originally modelled on the human brain. These networks consist of layers of nodes, one input layer, one or more hidden layers, and one output layer; each node in the network is connected to others with a certain weight and threshold \cite{ibm_what_2021}. There are many different types of neural networks, but they are typically large and complex. SVMs are smaller and less complex than neural networks; they work by separating classes using planes with varying functions; this technique is generally good on small but complex data \cite{geron_hands-machine_2017}. 

\begin{figure*}[htbp]
     \centering
     \begin{subfigure}[b]{0.45\textwidth}
         \centering
         \includegraphics[width=\textwidth]{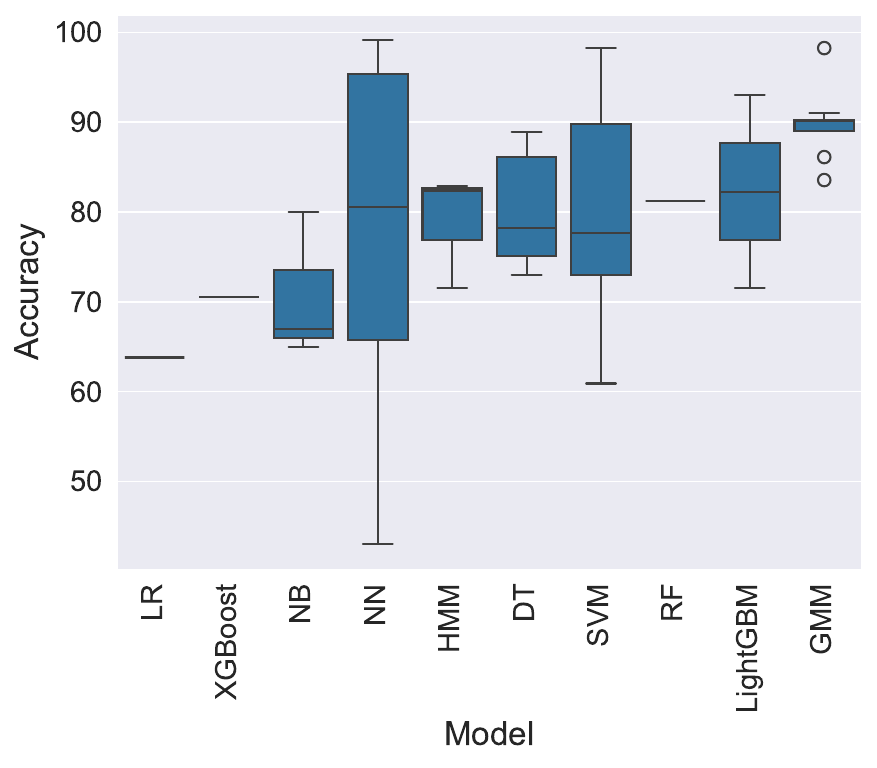}
    \caption{The accuracy achieved in the binary classifiers using different classification methods}
    \label{fig:BinaryMethodResultsComparison}
     \end{subfigure}
     \hfill
     \begin{subfigure}[b]{0.45\textwidth}
         \centering
        \includegraphics[width=\textwidth]{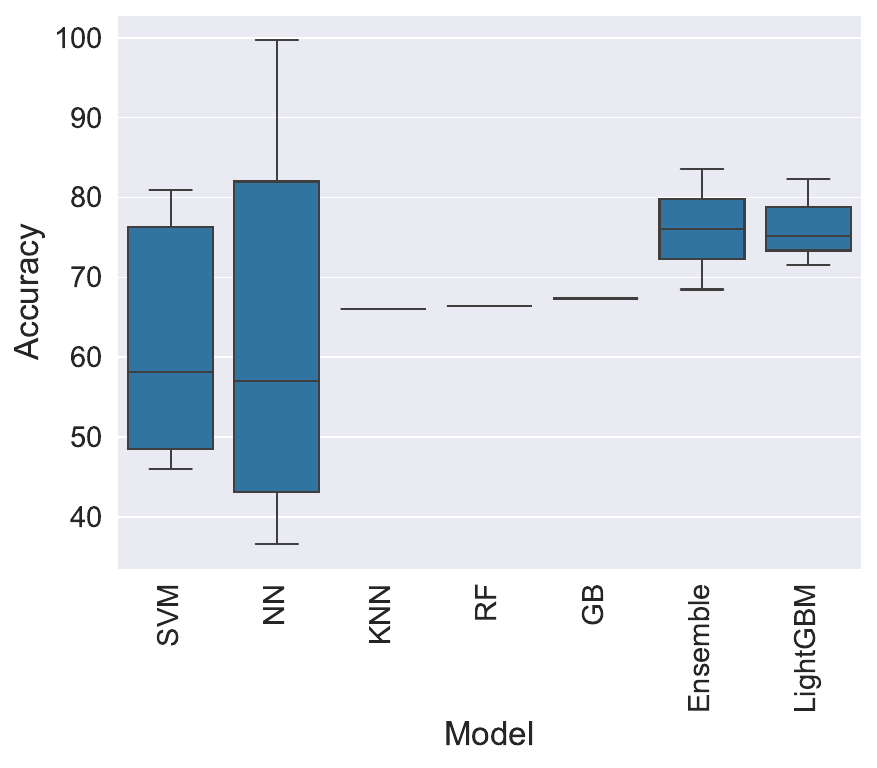}
    \caption{The accuracy achieved in the multi-class classifiers using different classification methods}
    \label{fig:MultiMethodResultsComparison}
     \end{subfigure}
     
        \caption{The results of different classification methods from binary and multi-class tasks.}
        \label{fig:ClassificationMethodsResults}
\end{figure*}

Figure \ref{fig:BinaryMethodResultsComparison} shows the accuracy achieved by each classification method used in articles for binary classification. While neural networks are the most commonly used algorithm, their performance varies greatly, having both the lowest and highest accuracy. Gaussian mixture models have the highest average performance. This figure shows that there is no one model that outperforms the rest of the models.

Figure \ref{fig:MultiMethodResultsComparison} shows the accuracy achieved by each classification method used in articles for multi-class classification. For multi-class classification, all methods performed similarly in terms of accuracy, with neural networks again varying the most, having both the highest and lowest accuracy. k-nn, random forest and gradient boosting all have very similar performance. However, each of these methods is only used once, and so it is difficult to say with any significance that they performed better than the other algorithms. LightGBM has the best average accuracy with ensamble methods close behind. Ensemble methods involve combining multiple classification algorithms with the hopes of enhancing the strengths of each method, this seems to be an effective method from the results seen in Figure \ref{fig:MultiMethodResultsComparison}. 

\begin{figure*}[htbp]
     \centering
     \begin{subfigure}[b]{0.45\textwidth}
         \centering
         \includegraphics[width=\textwidth]{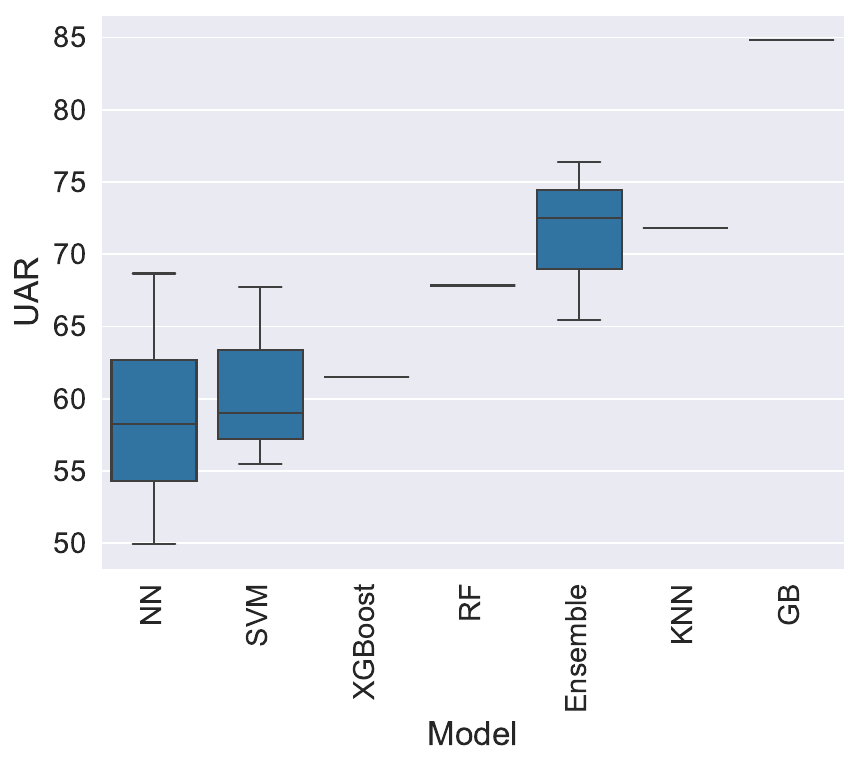}
    \caption{The UAR achieved in the IEEE FEMH Voice Data Challenge 2018 evaluated on cross-validation using different classification methods}
    \label{fig:ChallengeCrossVal}
     \end{subfigure}
     \hfill
     \begin{subfigure}[b]{0.45\textwidth}
         \centering
        \includegraphics[width=\textwidth]{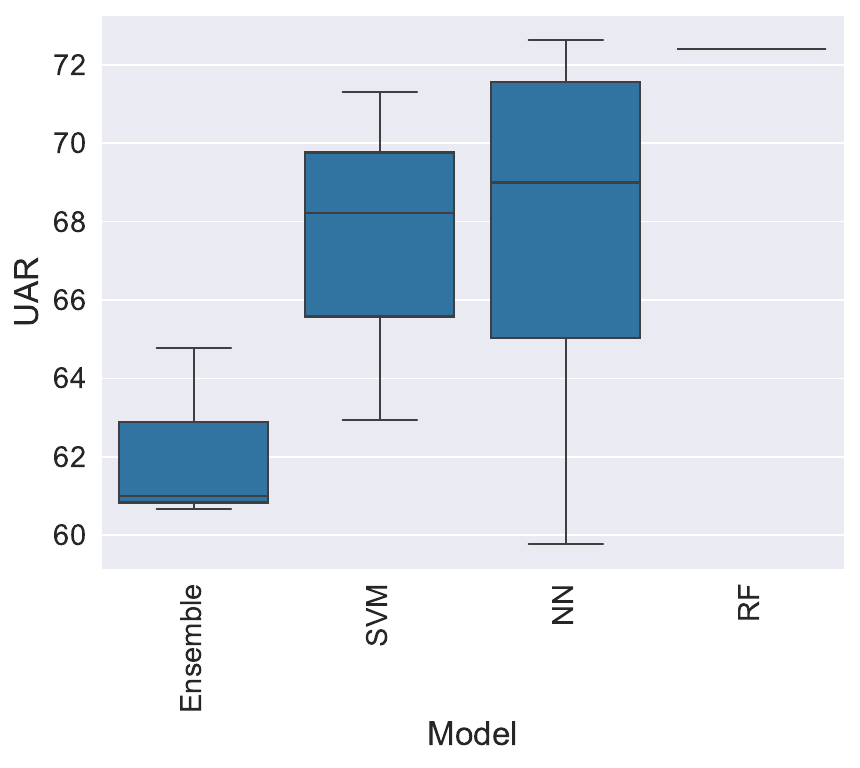}
    \caption{The UAR achieved in the IEEE FEMH Voice Data Challenge 2018 evaluated on a holdout test set using different classification methods}
    \label{fig:ChallengeHoldout}
     \end{subfigure}
     
        \caption{The UAR found in the articles submitted to the IEEE FEMH Voice Data Challenge 2018 for different classification methods \cite{grzywalski_parameterization_2018, ju_multi-representation_2018, islam_transfer_2018, arias-londono_byovoz_2018, pham_diagnosing_2018, chuang_dnn-based_2018, bhat_femh_2018, ramalingam_ieee_2018, pishgar_pathological_2018, degila_ucd_2018}.}
        \label{fig:ChallengeMethodsResults}
\end{figure*}

While these accuracy metrics provide valuable insights, it is important to note that the dataset used plays a crucial role in the performance of each technique. Therefore, comparing articles that use different datasets is challenging. However, the IEEE FEMH Voice Data Challenge 2018 provides a unique opportunity for direct comparison, as participants were given the same dataset. Participants were provided with 200 samples to be used in training (40 neoplasm, 60 phonotrauma, 50 vocal palsy, and 50 healthy) and models were then tested on an undisclosed test set of 400 samples \cite{bhat_femh_2018}. Figure \ref{fig:ChallengeMethodsResults} shows the UAR found using different methods in the challenge articles. UAR is used in this case as it is the most commonly reported metric amongst the challenge articles. Ten articles were submitted to this challenge \cite{grzywalski_parameterization_2018, ju_multi-representation_2018, islam_transfer_2018, arias-londono_byovoz_2018, pham_diagnosing_2018, chuang_dnn-based_2018, bhat_femh_2018, ramalingam_ieee_2018, pishgar_pathological_2018, degila_ucd_2018}. 

 Only six of the challenge articles report results on a holdout test set \cite{arias-londono_byovoz_2018, degila_ucd_2018, grzywalski_parameterization_2018, islam_transfer_2018, ju_multi-representation_2018, ramalingam_ieee_2018}. Of the methods reported on a holdout test set (Figure \ref{fig:ChallengeHoldout}), the random forest performs best, although there is only one instance. Neural networks have the second-best average performance but a very wide range, having the single best and worst performance. When evaluated on cross-validation, gradient boosting has a much higher accuracy than the other methods presented. However, this is a single result. Neural networks have the worst average performance, although their performance is the most wide-ranging of all the algorithms. This is a surprising result considering the performance of neural networks when evaluated on a holdout test set. Ensemble methods do well when evaluated using cross-validation but very poorly when evaluated on a holdout test set, this is possibly due to overfitting.

\subsection{RQ2 - Feature Extraction}
In this section, we discuss Research Question 2: What features of speech can be used to identify pathological speech, including throat cancer?

We present the features extracted from the examined articles and discuss how they perform. An average of 1.5 input features were used in each article with \citeauthor{kim_convolutional_2020} \cite{kim_convolutional_2020} using the most input features (raw signal, MFCC, acoustic features, spectral features). 

\begin{figure}[htbp]
    \centering
    \includegraphics[width=1\linewidth]{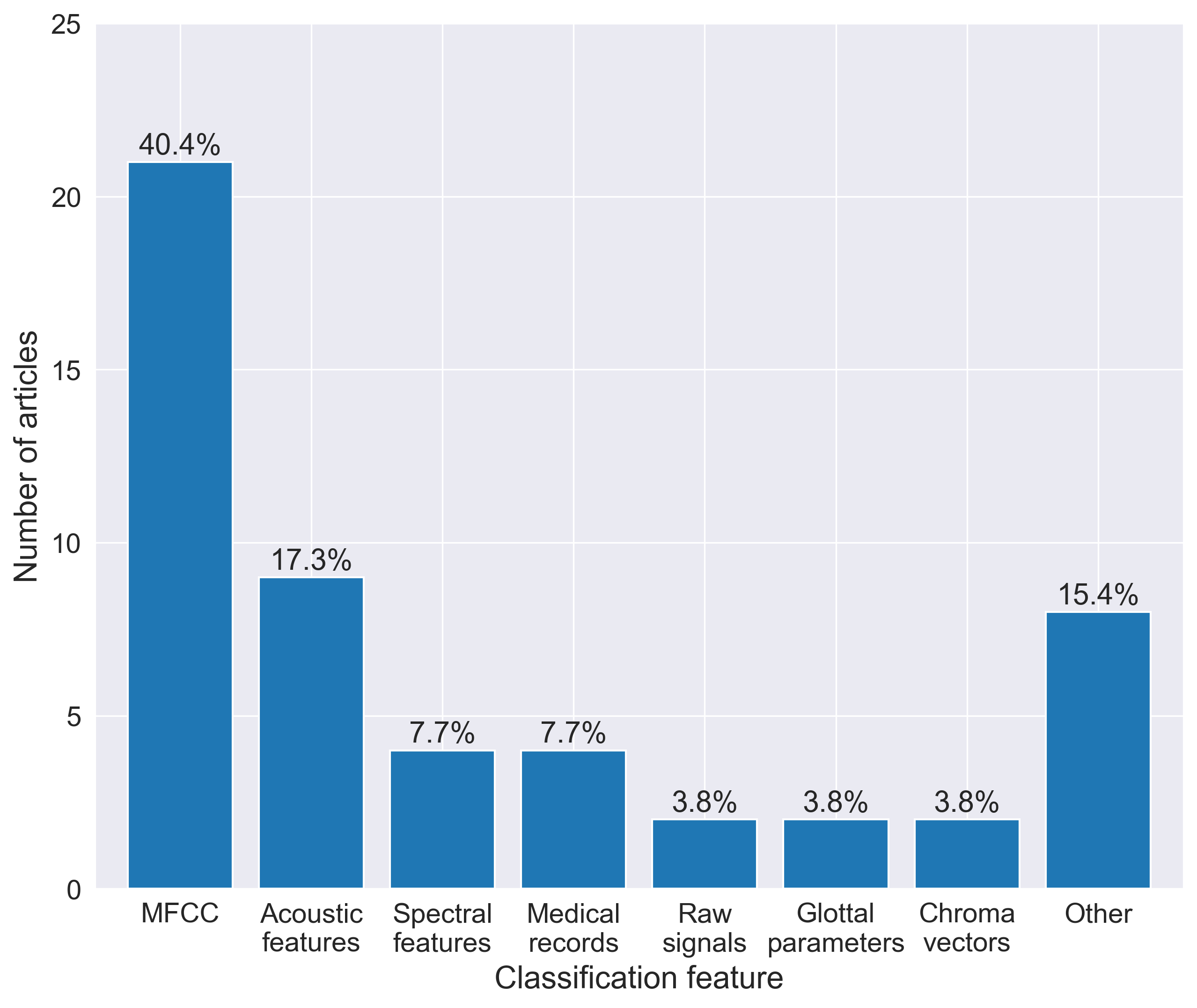}
    \caption{The features used across all articles found in this review. Any feature used only once was put into the other category for ease of visualisation. MFCC - Mel-frequency cepstral coefficients.}
    \label{fig:FeatureExtractionCount}
\end{figure}

In Figure \ref{fig:FeatureExtractionCount}, we show the number of each feature extraction method used in the articles. For clarity, methods used only once are grouped into the 'other' category. Some articles combined features as input into classification systems although this is not shown in Figure \ref{fig:FeatureExtractionCount}.

The most commonly used feature is Mel-frequence cepstral coefficients (MFCC). MFCCs represent the spectrum of an audio signal and are often treated as images when being input into classification systems. MFCCs are calculated using three steps \cite{godino-llorente_automatic_2004}: 
\begin{enumerate}
    \item Transforming the signal from the time to frequency domain using the short-time Fast Fourier Transform
    \item Finding the energy spectrum of each frame and calculating the energy in each mel window
    \item Taking logarithms and applying the cosine transform
\end{enumerate}

The second most common feature extracted was acoustic features which are extracted directly from the audio signal without prior transformation. Common acoustic features include jitter, shimmer, and harmonic features \cite{boersma_praat_2001, eyben_opensmile_2010}. Articles that use these two features often source work in other domains as a reason for their use.

\begin{figure*}[htbp]
     \centering
     \begin{subfigure}[b]{0.45\textwidth}
         \centering
         \includegraphics[width=\textwidth]{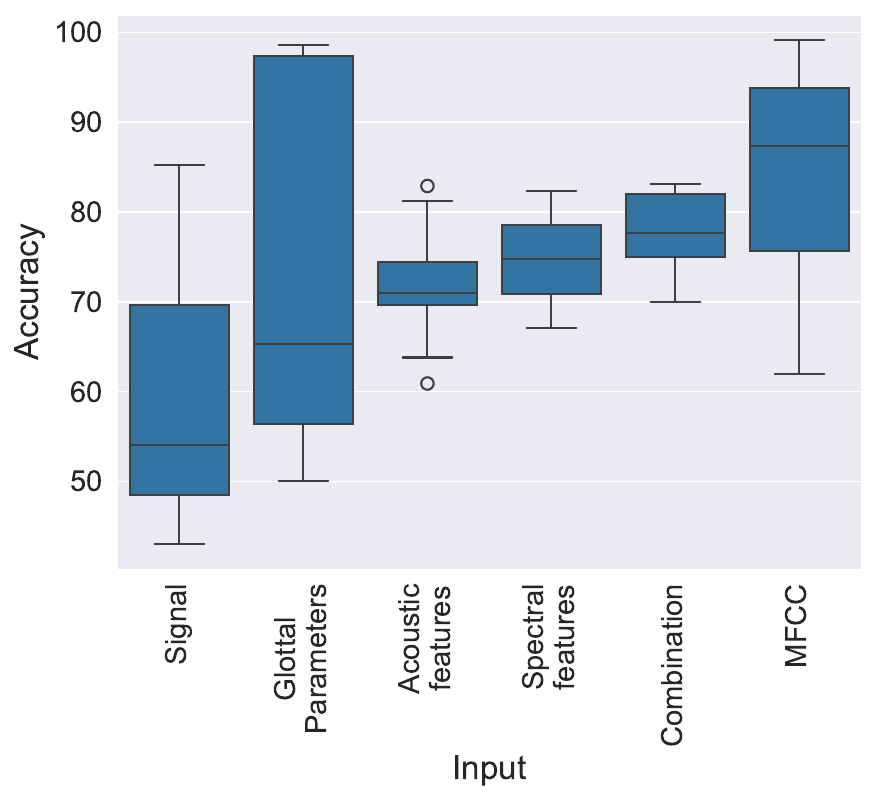}
    \caption{The accuracy achieved in the binary classifiers using different features}
    \label{fig:BinaryFeatureResultsComparison}
     \end{subfigure}
     \hfill
     \begin{subfigure}[b]{0.45\textwidth}
         \centering
        \includegraphics[width=\textwidth]{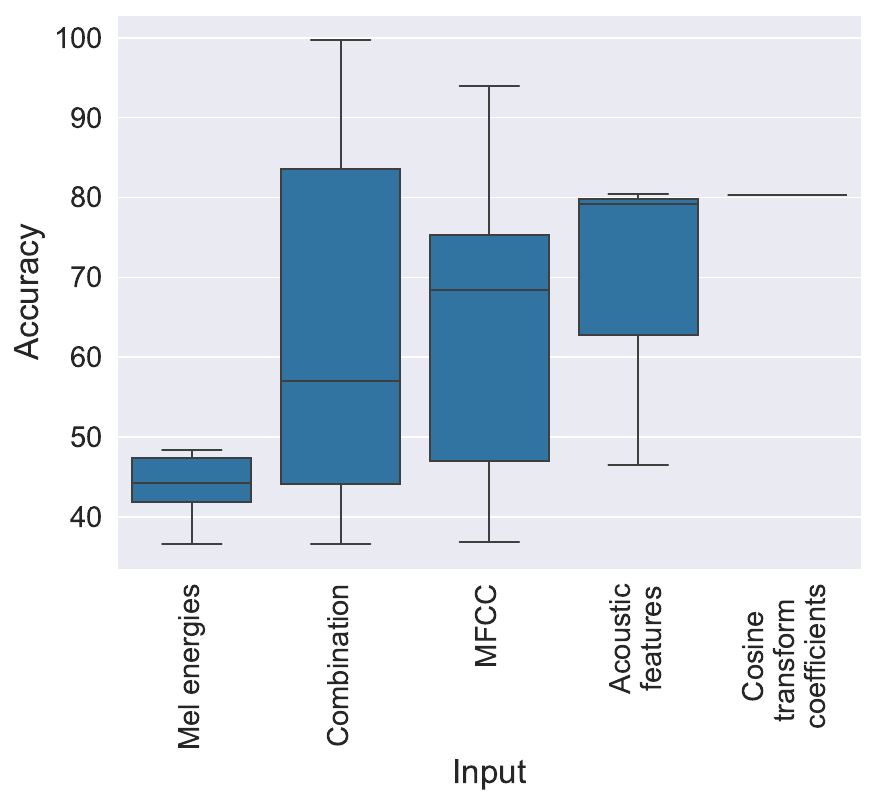}
    \caption{The accuracy achieved in the multi-class classifiers using different features}
    \label{fig:MultiFeatureResultsComparison}
     \end{subfigure}
     
        \caption{The results of different feature extraction methods from binary and multi-class tasks.}
        \label{fig:FeatureExtractionResults}
\end{figure*}

Figure \ref{fig:FeatureExtractionResults} shows the results obtained from methods using different features. Figure \ref{fig:BinaryFeatureResultsComparison} shows the results obtained for binary classification. MFCC has the highest average performance, with the raw signal having the worst average performance. The poor performance of using a raw signal as input may be due to the large dimensionality of an audio signal, without extracting features, there will be lots of noise and redundant data input into the system. Only two articles use raw signals as input to the system \cite{kim_convolutional_2020, wang_ai_2024}. Glottal parameters have the broadest range of results, although these results come from only two articles with shared authors \cite{ben_aicha_cancer_2016, ezzine_towards_2016}. 

Figure \ref{fig:MultiFeatureResultsComparison} shows the results obtained for multi-class classification. Cosine transform coefficients have the best average performance although there is only a single instance of these being used as system input. Mel energies have the lowest average and lowest single performance, although a single article obtains these results \cite{chen_classification_2023}. The single best result is obtained by \citeauthor{song_enhancing_2023} \cite{song_enhancing_2023} using a combination of features, this combination contains MFCC and acoustic features, as well as demographic features. 

\begin{figure*}[htbp]
     \centering
     \begin{subfigure}[b]{0.45\textwidth}
         \centering
         \includegraphics[width=\textwidth]{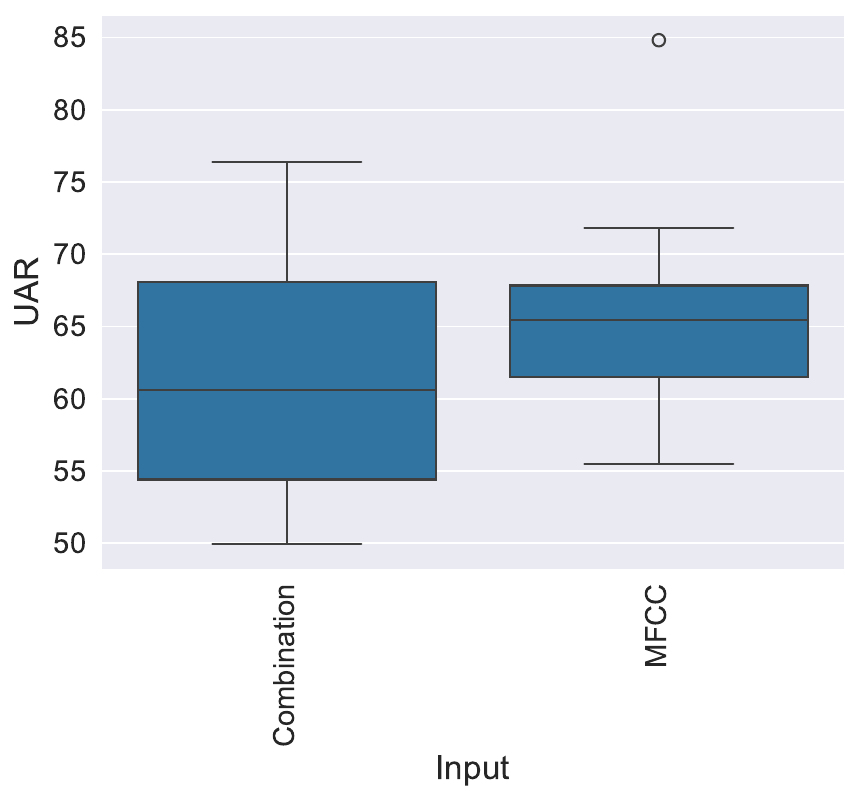}
    \caption{The UAR achieved in the IEEE FEMH Voice Data Challenge 2018 evaluated on cross-validation using different system inputs}
    \label{fig:ChallengeCrossValFeatures}
     \end{subfigure}
     \hfill
     \begin{subfigure}[b]{0.45\textwidth}
         \centering
        \includegraphics[width=\textwidth]{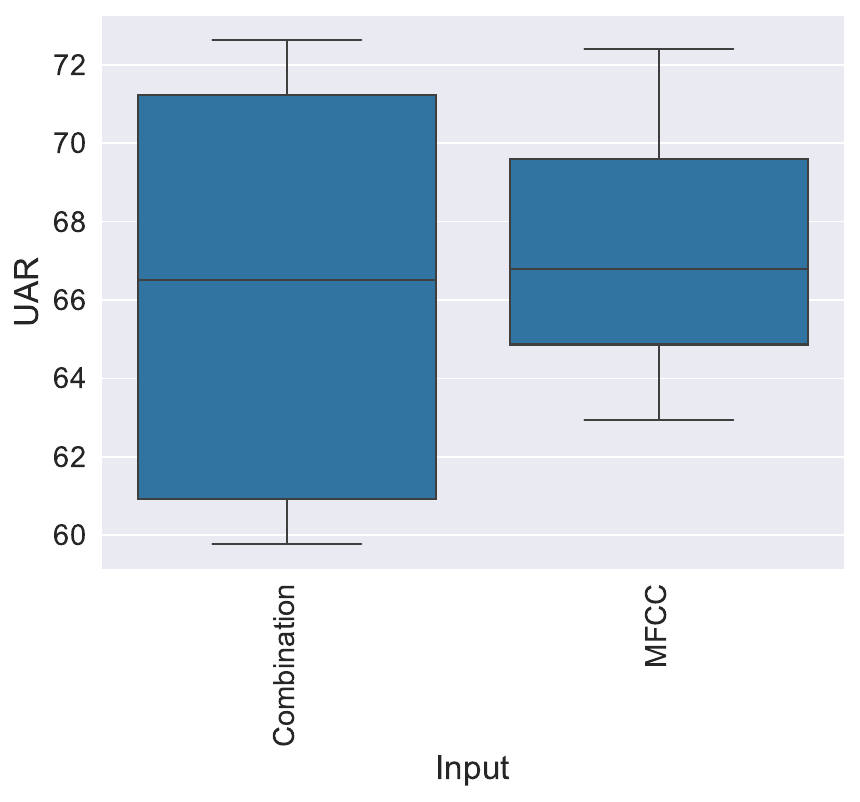}
    \caption{The UAR achieved in the IEEE FEMH Voice Data Challenge 2018 evaluated on a holdout test set using different system inputs}
    \label{fig:ChallengeHoldoutFeatures}
     \end{subfigure}
     
        \caption{The UAR found in the articles submitted to the IEEE FEMH Voice Data Challenge 2018 for different system inputs \cite{grzywalski_parameterization_2018, ju_multi-representation_2018, islam_transfer_2018, arias-londono_byovoz_2018, pham_diagnosing_2018, chuang_dnn-based_2018, bhat_femh_2018, ramalingam_ieee_2018, pishgar_pathological_2018, degila_ucd_2018}.}
        \label{fig:ChallengeFeatureResults}
\end{figure*}

As stated in the previous section, comparing the results of articles that use different datasets is hard. However, we can compare the results of the articles submitted to the IEEE FEMH Voice Data Challenge 2018. Figure \ref{fig:ChallengeFeatureResults} shows the results obtained based on the features used, this is reported as UAR as this was the metric most widely reported in the challenge articles. In the challenge articles, only two types of features were used: MFCC and a combination of different features. This combination generally included MFCC, acoustic features, and spectral features. Figure \ref{fig:ChallengeCrossValFeatures} shows the results of articles on cross-validation.

\subsection{RQ3 - Strengths,  Weaknesses and Recomendations}
\begin{figure}[]
    \centering
    \includegraphics[width=1\linewidth]{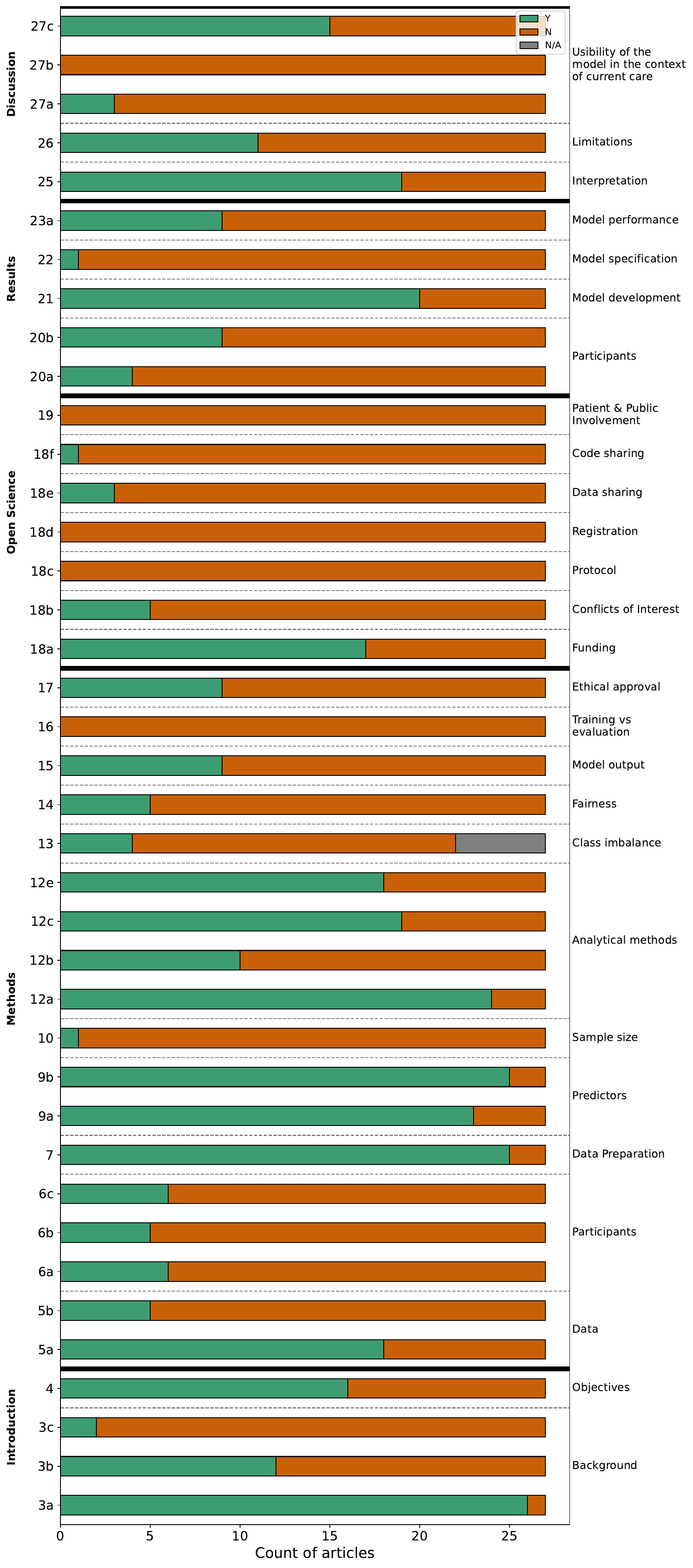}
    \caption{The results of comparing the articles to the TRIPOD-AI checklist. Any points that were not applicable to any of the articles in this study were removed. The labels on the left show the sections of the TRIPOD-AI checklist that each item was listed under, and the labels on the right show the subheading for each of the items.}
    \label{fig:TRIPOD_results}
\end{figure}

In this section, we address Research Question 3: What are the strengths of the existing research, and what issues need to be addressed in future work? To answer this, we compare each paper against the Transparent Reporting of a multivariable prediction model for Individual Prognosis Or Diagnosis (TRIPOD) AI checklist \cite{collins_tripodai_2024}. The TRIPOD-AI checklist sets out guidelines for the reporting of studies using AI or ML for the diagnosis or classification of medical conditions. We do acknowledge that this study was published in January 2024 and, as such, was not available to the authors of many of the studies found in this search. However, we still feel this is a good baseline against which the articles can be compared. TRIPOD-AI is split into 27 sections, each with several subsections totalling 50 points to be found in a study, although some of these points are not applicable to the studies in this search. Figure \ref{fig:TRIPOD_results} shows the number of studies that included each of the points; for clarity, any points that were found to be not applicable to any of the studies have been removed. This figure shows the individual points that are defined in the TRIPOD-AI checklist, as well as the sections and subsections that they belong to. Table \ref{tab:TRIPODPaperResults} shows the percentage of points each article achieved split by sections; we excluded sections that were not applicable for each article. We summarise our findings in this section split into key weaknesses and key strengths.

\begin{table*}[htbp]
\centering
\begin{tabular}{l|l|l|l|l|l}
& Introduction & Methods & Open Science & Results & Discussion \\ \hline
\citeauthor{gavidia-ceballos_direct_1996} \cite{gavidia-ceballos_direct_1996} & 25\% & 28\% & 0\% & 0\% & 20\% \\
\citeauthor{godino-llorente_automatic_2004} \cite{godino-llorente_automatic_2004} & 25\% & 44\% & 0\% & 40\% & 0\% \\
\citeauthor{verikas_towards_2010} \cite{verikas_towards_2010} & 50\% & 39\% & 29\% & 40\% & 40\% \\
\citeauthor{ben_aicha_cancer_2016} \cite{ben_aicha_cancer_2016} & 25\% & 17\% & 0\% & 0\% & 20\% \\
\citeauthor{ezzine_towards_2016} \cite{ezzine_towards_2016} & 50\% & 29\% & 14\% & 20\% & 20\% \\
\citeauthor{arias-londono_byovoz_2018} \cite{arias-londono_byovoz_2018} & 75\% & 39\% & 14\% & 20\% & 40\% \\
\citeauthor{bhat_femh_2018} \cite{bhat_femh_2018} & 25\% & 28\% & 0\% & 20\% & 0\% \\
\citeauthor{chuang_dnn-based_2018} \cite{chuang_dnn-based_2018} & 50\% & 33\% & 14\% & 20\% & 20\% \\
\citeauthor{degila_ucd_2018} \cite{degila_ucd_2018} & 75\% & 33\% & 0\% & 20\% & 60\% \\
\citeauthor{grzywalski_parameterization_2018} \cite{grzywalski_parameterization_2018} & 50\% & 47\% & 0\% & 20\% & 0\% \\
\citeauthor{islam_transfer_2018} \cite{islam_transfer_2018} & 75\% & 44\% & 14\% & 20\% & 0\% \\
\citeauthor{ju_multi-representation_2018} \cite{ju_multi-representation_2018} & 50\% & 44\% & 0\% & 20\% & 40\% \\
\citeauthor{pham_diagnosing_2018} \cite{pham_diagnosing_2018} & 75\% & 39\% & 14\% & 20\% & 40\% \\
\citeauthor{pishgar_pathological_2018} \cite{pishgar_pathological_2018} & 75\% & 33\% & 14\% & 20\% & 20\% \\
\citeauthor{ramalingam_ieee_2018} \cite{ramalingam_ieee_2018} & 0\% & 22\% & 14\% & 0\% & 20\% \\
\citeauthor{fang_combining_2019} \cite{fang_combining_2019} & 50\% & 44\% & 0\% & 40\% & 60\% \\
\citeauthor{fang_detection_2019} \cite{fang_detection_2019} & 25\% & 50\% & 14\% & 60\% & 80\% \\
\citeauthor{kim_convolutional_2020} \cite{kim_convolutional_2020} & 50\% & 71\% & 29\% & 60\% & 40\% \\
\citeauthor{miliaresi_combining_2021} \cite{miliaresi_combining_2021} & 25\% & 41\% & 14\% & 20\% & 40\% \\
\citeauthor{kwon_diagnosis_2022} \cite{kwon_diagnosis_2022} & 100\% & 67\% & 14\% & 20\% & 60\% \\
\citeauthor{wang_detection_2022} \cite{wang_detection_2022} & 50\% & 82\% & 0\% & 60\% & 40\% \\
\citeauthor{chen_classification_2023} \cite{chen_classification_2023} & 25\% & 56\% & 14\% & 60\% & 20\% \\
\citeauthor{paterson_pipeline_2023} \cite{paterson_pipeline_2023} & 75\% & 44\% & 43\% & 60\% & 60\% \\
\citeauthor{song_enhancing_2023} \cite{song_enhancing_2023} & 75\% & 22\% & 14\% & 60\% & 20\% \\
\citeauthor{zaim_accuracy_2023} \cite{zaim_accuracy_2023} & 75\% & 72\% & 29\% & 60\% & 60\% \\
\citeauthor{kim_classification_2024} \cite{kim_classification_2024} & 75\% & 72\% & 29\% & 60\% & 60\% \\
\citeauthor{wang_ai_2024} \cite{wang_ai_2024} & 50\% & 50\% & 43\% & 20\% & 80\%
\end{tabular}
\caption{A table showing the percentage of points in the TRIPOD-AI checklist hit by each article of those applicable to that article.}
\label{tab:TRIPODPaperResults}
\end{table*}

\subsubsection{Key Weaknesses}
When comparing the papers to TRIPOD-AI we observed notable discrepancies between journal papers (average coverage of 18 points) and conference papers (average coverage of 12 points). We speculate that this disparity may arise from the typically stricter page limits imposed on conference papers compared to journal articles.

A critical area of weakness identified through TRIPOD-AI assessment is the lack of adherence to open science (points 18a-18f, and 19). Shockingly, only one of the reviewed articles has made its code available to the general public \cite{paterson_pipeline_2023}; the lack of code sharing severely reduces the ability of other researchers to reproduce their results or validate their methods using external data. Furthermore, only three articles utilize open-source datasets \cite{paterson_pipeline_2023, ezzine_towards_2016, wang_ai_2024}. Additionally, many articles fail to report conflicts of interest (point 20) and provide details on study protocol (point 22) or registration (point 22). While some articles mention data availability upon request, this practice is considered inadequate in promoting transparency \cite{tedersoo_data_2021}.

Another significant issue is lack of information about the participants within the datasets, with point 6a being covered by six articles \cite{kim_convolutional_2020, wang_detection_2022, kwon_diagnosis_2022, chen_classification_2023, zaim_accuracy_2023, kim_classification_2024}, 6b being covered by five articles \cite{kim_convolutional_2020, wang_detection_2022, kwon_diagnosis_2022, chen_classification_2023, kim_classification_2024}, and 6c being covered by six articles \cite{kim_convolutional_2020, wang_detection_2022, kwon_diagnosis_2022, chen_classification_2023, kim_classification_2024, wang_ai_2024}. These points cover a discussion of the study setting, eligibility criteria for participants, and details of any treatments received. We do acknowledge that this information is not always available when using data provided by a third party; however, we feel that this calls for better data descriptions to be provided when data is disseminated. We also believe that it is important to acknowledge the missing information about the data when discussing the limitations of the work. We have found that the articles, in general, are hard to compare due to the diversity in data sources and evaluation metrics. This is made worse by the lack of open data sharing, as subsequent studies cannot use the same data sources and be able to compare directly to previous studies. 

Another area where articles fall short compared to the TRIPOD-AI checklist is the lack of discussion on the practical application of AI systems in real-world medical settings (points 27a-27c). While half of the articles discuss steps for further research, only one addresses the impact of poor-quality input data, and none of the articles discuss the user inputs to a possible system. This gap in research indicates a need for more comprehensive studies that consider the practical implementation of AI systems in real-world medical settings, although this may be due to the relative newness of this area of study. 

Additionally, the dependence on small datasets is concerning.  As stated in Section \ref{sec:Overview}, the largest reported dataset is that of \citeauthor{ezzine_towards_2016} \cite{ezzine_towards_2016}, although we believe that this is an outlier due to omitted preprocessing steps, meaning that this large dataset does not represent the number of participants. The next largest dataset is that of \citeauthor{wang_ai_2024} \cite{wang_ai_2024}, containing 2000 patients. Excluding \citeauthor{ezzine_towards_2016} \cite{ezzine_towards_2016} and \citeauthor{ben_aicha_cancer_2016} \cite{ben_aicha_cancer_2016} due to unclear dataset size, the mean size of the datasets across the articles is 306 participants. We also note that only three of the included articles perform any external validation of their models \cite{kwon_diagnosis_2022, fang_detection_2019, wang_ai_2024}; combining this with the small sample size puts them at high risk of bias and overfitting \cite{wolff_probast_2019}.

\subsubsection{Key Strengths}
On the other hand, strong areas identified through TRIPOD-AI analysis include robust reporting on data preparation and predictor definition (feature extraction techniques). Most papers include the steps for feature extraction or the equations used. We also find that the description of the healthcare context is generally strong, with clear motives for the work.

One of the key strengths in this area of study is the impressive results obtained by some of the articles. The binary classification accuracy ranges from 81-99\%, showcasing the potential of the research. While the performance of the multi-class articles is harder to compare due to varying class and sample sizes, the variation in classification methods and feature extraction techniques demonstrates the promising future and potential for novel exploration in this field.

\subsubsection{Recommendations}
From these results, we can formulate some recommendations for future work in this area. It is clear that open science needs to be improved with future work, starting with code sharing; by providing a link to code availability within the article, other researchers are better able to understand all steps taken in the work and use it to improve their research. We believe that code sharing has very few disadvantages to authors. While we are aware of the limitations in making data available due to concerns around patient privacy and anonymity, we feel that better data descriptions and external validation on a publically available data source could allow for a better understanding of how these studies may be implemented in clinical practice and, alongside code sharing, allow for the validation of results. With the recent development of TRIPOD-AI, authors should attempt to adhere to the checklist where possible and acknowledge where they cannot provide information (for example, when information about a dataset is missing).

\section{Conclusion}
Within this scoping literature review, we identified 27 articles describing the classification of vocal pathologies using machine learning and speech recordings. Of these, 13 performed multi-class classification, 12 focused on binary classification, and two did both binary and multi-class classification. Our findings effectively addressed the research questions outlined in Section \ref{sec:RQs}. 

Neural networks emerged as the most commonly employed classification algorithm, although other algorithms demonstrated comparable success in classifying patients. The most common features used were MFCCs, although a wide variety of features were explored without a clear superiority observed among them. The results found in these works show that this is a promising area and that the use of AI in the detection of speech pathologies should be further investigated. We discuss the weaknesses in this area, particularly the lack of open science and the lack of publicly available code and datasets. The absence of publicly available code or datasets prevents the reproducibility and extension of findings across these articles. We also make recommendations for future research in this area to promote better open science and clarity of reporting. In future work, we will address this shortcoming, with the aim to create public code repositories, hence enabling external validation and facilitating improvements in subsequent studies. 

\section*{Acknowledgements}
This research was funded in part by the UKRI Engineering and Physical Sciences Research Council (EPSRC) [EP/S024336/1].

\clearpage
\bibliographystyle{IEEEtranN}
\bibliography{references}






\end{document}